\documentclass[runningheads]{llncs}

\usepackage{graphicx}
\usepackage{amsmath}
\usepackage{amssymb}
\usepackage{booktabs}
\usepackage{algorithmic}
\usepackage{comment}
\usepackage{hyperref}
\usepackage{subcaption}

\begin{document}

\title{Generalized Probabilistic U-Net for medical image segementation}

\author{Ishaan Bhat\thanks{Corresponding Author}\inst{1}\orcidID{0000-0002-4398-018X} \and Josien P.W. Pluim\inst{2, 1}\orcidID{0000-0001-7327-9178} \and Hugo J. Kuijf\inst{1}\orcidID{0000-0001-6997-9059}}

\authorrunning{I. Bhat et al.}
\institute{Image Sciences Institute, University Medical Center Utrecht, Heidelberglaan 100, 3584 CX Utrecht, The Netherlands\\ \email{\{i.r.bhat, h.kuijf\}@umcutrecht.nl} \and Department of Biomedical Engineering, Eindhoven University of Technology, Groene Loper 3, 5612 AE Eindhoven, The Netherlands\\ \email{j.pluim@tue.nl}}

\maketitle              

\begin{abstract}
We propose the Generalized Probabilistic U-Net, which extends the Probabilistic U-Net\cite{kohl_probabilistic_2018} by allowing more general forms of the Gaussian distribution as the latent space distribution that can better approximate the uncertainty in the reference segmentations. We study the effect the choice of latent space distribution has on capturing the uncertainty in the reference segmentations using the LIDC-IDRI dataset. We show that the choice of distribution affects the sample diversity of the predictions and their overlap with respect to the reference segmentations. For the LIDC-IDRI dataset, we show that using a mixture of Gaussians results in a statistically significant improvement in the generalized energy distance (GED) metric with respect to the standard Probabilistic U-Net. We have made our implementation available at https://github.com/ishaanb92/GeneralizedProbabilisticUNet
\end{abstract}

\section{Introduction}
Image segmentation may be posed as a supervised classification task with a deep learning system trained using manually created expert labels producing a segmentation map by estimating per-voxel class probabilities. A shortcoming of standard deep learning approaches is that they produce point estimate predictions and not an output distribution from which multiple plausible predictions can be sampled~\cite{kendall_what_2017}. This disadvantage is highlighted when multiple annotations per image are available, which standard deep learning approaches cannot leverage. Inter-observer variability reflects the disagreement among experts and ambiguity present in the image, and ideally, a supervised learning approach needs to reflect these for unseen test cases~\cite{jungo_effect_2018}.

Recent works such as the Probabilistic U-Net~\cite{kohl_probabilistic_2018} and PHISeg framework~\cite{baumgartner_phiseg_2019} were developed to handle segmentation of ambiguous images by using multiple annotations per image during training. They used variational inference~\cite{wainwright_graphical_2007} to produce multiple spatially coherent predictions per image by sampling from a distribution over a (learned) latent space. Key to their approach is modelling the latent space distribution as an axis-aligned Gaussian, i.e. a multivariate Gaussian distribution with a diagonal covariance matrix. It has been hypothesized that the choice of a simple distribution reduces sample diversity and a more expressive distribution might be required for producing more realistic segmentations~\cite{liu_uncertainty_2020}, \cite{valiuddin_improving_2021}. 

In this paper, we extend the Probabilistic U-Net framework by using more general forms of the Gaussian distribution to study the effect the choice of the latent space distribution has on capturing the uncertainty in medical images.

\section{Related Work}
\label{sec:rel_work}
In \cite{kohl_probabilistic_2018} the authors combine the conditional variational autoencoder (cVAE)~\cite{sohn_learning_2015} framework with the popular U-Net~\cite{ronneberger_u-net_2015} architecture to create the Probabilistic U-Net. Different variants of the prediction are computed by sampling from the (learned) latent space and combining this sample with the highest resolution feature map of the U-Net. In \cite{hu_supervised_2019} the authors extend the Probabilistic U-Net to capture model uncertainty by using variational dropout~\cite{kingma_variational_2015} over the model weights. The PHISegNet~\cite{baumgartner_phiseg_2019} and Hierarchical Probabilistic U-Net~\cite{kohl_hierarchical_2019} show a further improvement in sample diversity by using a series of hierarchical latent spaces from which samples are combined with the U-Net feature maps at different resolutions. The Probabilistic U-Net and its hierarchical variants use an axis-aligned Gaussian distribution to model the distribution over the latent space. However, techniques such as normalizing flows~\cite{rezende_variational_2016}, that convert simple distributions into complex ones via invertible transformations, have been shown to be a promising alternative to model the latent space distribution. The cFlowNet~\cite{liu_uncertainty_2020} uses normalizing flows to model the latent space distribution to produce plausible and diverse outputs.

\begin{figure*}[htb]
\centering
\includegraphics[width=1.0\linewidth]{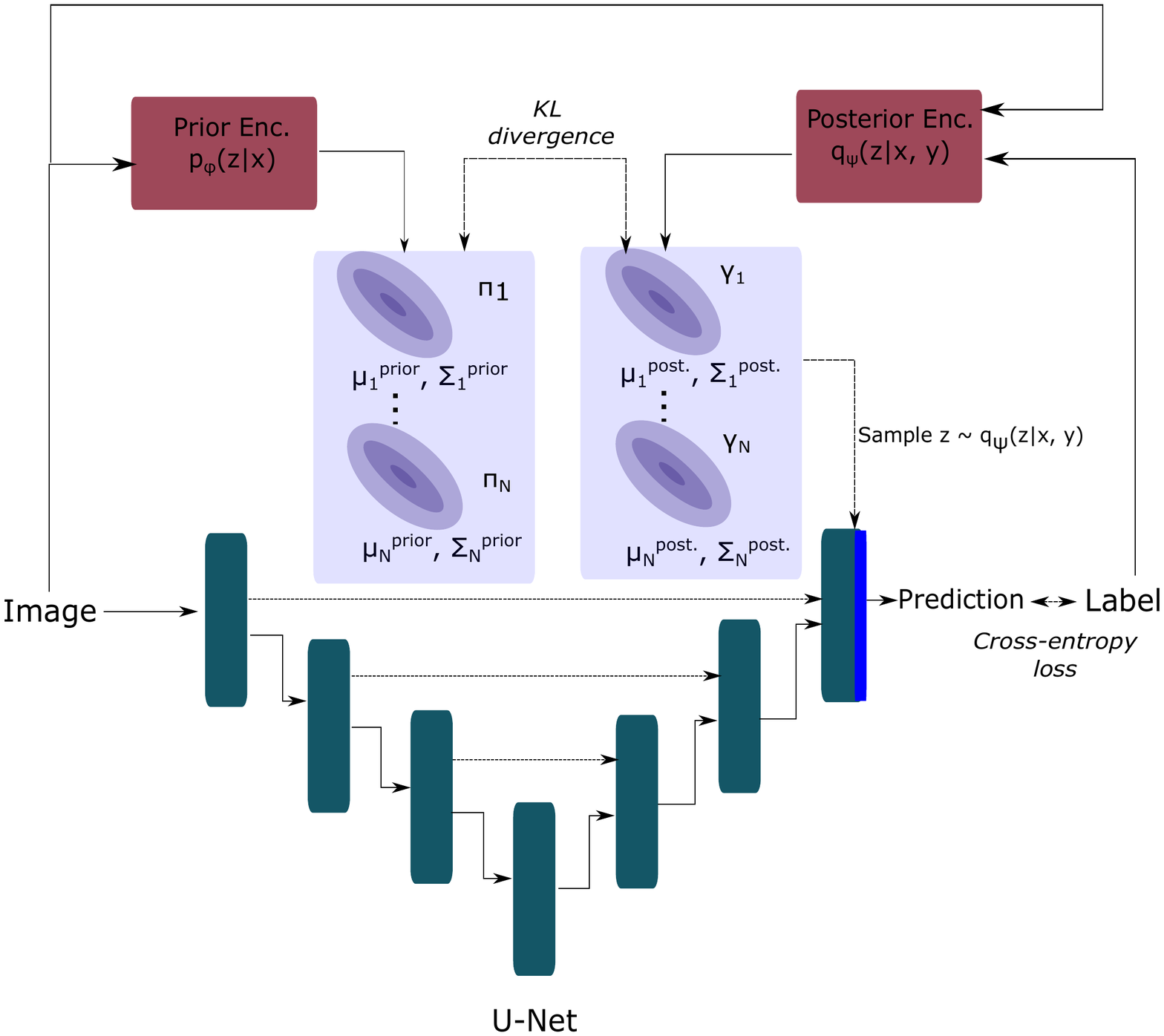}
\caption{Generalized Probabilistic U-Net framework. During training, the model learns prior and posterior distribution parameters $\{\mu_i^{\mathrm{prior}}, \Sigma_i^{\mathrm{prior}}, \pi_{i}\}_{i=1}^N$ and $\{\mu_i^{\mathrm{post.}}, \Sigma_i^{\mathrm{post.}}, \gamma_{i}\}_{i=1}^N$. In the most general case, the prior and posterior distributions are modelled as a mixture of $N$ Gaussians. Similarly, different variants of the multivariate Gaussian distribution can be modelled by constructing the covariance matrix in the different ways described in Section \ref{sec:contribution}. By setting $N=1$ and restricting the covariance matrix to be diagonal, we recover the original Probabilistic U-Net. During inference, the posterior encoder is discarded, and different plausible outputs can be computed by sampling from the prior distribution and combining this sample with the last U-Net layer.}
\label{fig:punet}
\end{figure*}

\section{Contributions}
\label{sec:contribution}

In an image segmentation problem, we typically have an image $x \in \mathbb{R}^{H \times W \times C}$ and label $y \in \{0, 1\}^{H \times W}$ and our goal is to produce accurate and coherent samples from the distribution $p_{\theta}(y | x)$ using a deep neural network parameterized by $\theta$. To enable tractable sampling from $p_{\theta}(y|x)$, simple parameterizable distributions $p_{\phi}(z|x)$ and $q_{\psi}(z|x, y)$ are used as approximations for the true prior and posterior distributions over the latent space $z$ respectively. The distributional parameters for $q_{\psi}(z|x, y)$ and $p_{\phi}(z|x)$ are learned by neural (sub-)networks paramterized by $\psi$ and, $\phi$ respectively.  

To produce predictions that better approximate the uncertainty in the reference segmentations, we propose two extensions to the Probabilistic U-Net framework by using more general forms of the Gaussian distribution as choices for the latent space distributions. Our Generalized Probabilistic U-Net framework is shown in Figure \ref{fig:punet}.

\subsection{Full-covariance Gaussian}
For a matrix to be a valid covariance matrix, it must be  positive semidefinite. Since this constraint is difficult to impose while training a neural network, the covariance matrix $\Sigma$ is built using its Cholesky decomposition $L$~\cite{williams1996}:
\begin{equation*}
    \Sigma = LL^{T}
\end{equation*}

The matrix $L$ is a lower-triangular matrix with a positive-valued diagonal. By masking the upper-triangular section of the matrix and using the exponential operator to ensure positive values on the diagonal, $L$ can be directly computed by the neural network. Samples can be drawn from the full-covariance Gaussian using the reparameterization trick~\cite{kingma_introduction_2019}:
\begin{align*}
    z = \mu + L*\epsilon,\ \epsilon \sim \mathcal{N}(0, I)
\end{align*}

\subsection{Mixture of Gaussians}
Any arbitrary (continuous) distribution can be modelled by using a mixture of a sufficient number of Gaussians, with appropriate mixture weights~\cite{bishop2006}. Mixtures of Gaussians have been used to model posterior distributions in variational autoencoders for semi-supervised classification~\cite{Nalisnick2016ApproximateIF} and clustering~\cite{jiang_variational_2017}, \cite{kopf_mixture--experts_2021}.

For example, the posterior distribution\footnote{This subsection holds true for the cVAE prior distribution as well. The only difference is the dependence on the label, $y$, is dropped.} can be modelled as a mixture of Gaussians as follows:
\begin{equation*}
    q_{\psi}(z | x, y) = \sum_{i=1}^{N} \gamma_i \mathcal{N}(\mu_i(x, y), \Sigma_i(x, y))
\end{equation*}
The individual Gaussians in the mixture, $\mathcal{N}(\mu_i(x, y), \Sigma_i(x, y))$, are called the component distributions and the weights for the component distributions, $\{\gamma_i\}_{i=1}^N$, are the mixing coefficients. The individual Gaussians can be modelled using a diagonal or full covariance matrix. For the distribution to be valid, the mixing coefficients must be greater than or equal to $0$ and sum to $1$. Therefore, a categorical distribution can be used to define the mixture distribution.

A distribution chosen to model the posterior must support differentiable sampling so that optimization of the loss function may be performed via backpropagation. To sample from a mixture of Gaussians, first a component index is sampled from the categorical distribution defined by the mixture coefficients and then a value is sampled from the corresponding component (Gaussian) distribution. 
\begin{gather*}
    i \sim \mathrm{Cat}(N;\gamma) \\
    z \sim \mathcal{N}(\mu_i(x, y), \Sigma_i(x, y))
\end{gather*}
The second step in the sampling process is differentiable and supports the reparameterization trick, however, the first step i.e. sampling from a categorical distribution is not differentiable. To make sampling fully differentiable, we used the Gumbel-Softmax (GS) distribution\cite{jang_categorical_2017}, \cite{maddison_concrete_2017} to model the mixture distribution. The Gumbel-Softmax distribution is a continuous relaxation (controlled by the temperature parameter, $\tau$) of the discrete categorical distribution, that supports differentiable sampling via the reparameterization trick. To obtain a discrete component index, we perform Straight-Through(ST) sampling~\cite{jang_categorical_2017}, where we used the $\mathrm{argmax}$ operation in the forward pass and used the continuous relaxation in the backward pass while computing gradients. With this, we used the following two-step process to sample from the mixture of Gaussians:
\begin{gather*}
    i \sim \mathrm{GS}(N;\gamma, \tau) \\
    z \sim \mathcal{N}(\mu_i(x, y), \Sigma_i(x, y))
\end{gather*}

Unlike for a pair of Gaussians, the KL divergence for a pair of Gaussian mixtures, which is a part of the neural network loss function, does not have a closed-form expression. We estimated the KL divergence via Monte Carlo integration, which provides a good approximation~\cite{hershey2007}.

\section{Methodology}
\subsection{Data}

The LIDC-IDRI dataset~\cite{lidc2011} consists of $1018$ thoracic CT scans with lesion segmentations from four experts. Similar to \cite{kohl_probabilistic_2018}, \cite{baumgartner_phiseg_2019} we use a pre-processed version of the dataset, consisting of $128 \times 128$ patches containing lesions.

We obtained a total of $15,096$ patches. We used a $60:20:20$ split to get $9058,\,3019,$ and $3019$ patches for training, validation, and testing respectively. The intensity of the patches was scaled to $[0, 1]$ range and no data augmentation was used. We used the publicly available version of the pre-processed dataset available at https://github.com/stefanknegt/Probabilistic-Unet-Pytorch.

\subsection{Neural network training}

For all models we used the ADAM~\cite{kingma_adam:_2015} optimizer with a learning rate of $10^{-4}$. Training was stopped when the validation loss did not improve for more than 20 epochs, and the model parameters with the minimum loss were saved. 

We used the Tune~\cite{liaw2018tune} package to choose optimal hyperparameters via a random search strategy. Depending on the number of hyperparameters, the best performing model was chosen among $4-24$ instances. The hyperparameters for all models are shown in Appendix \ref{sec:hyperparams}.

To ensure comparability between models with different latent space distributions, we maintain the same number of layers for the U-Net, the prior encoder, and the posterior encoder. Furthermore, in accordance with the original Probabilistic U-Net architecture, the prior and posterior encoder have the same architecture as the U-Net encoder. Each convolution block in the model consisted of convolution with a $3 \times 3$ kernel, a ReLU nonlinearity and batch normalization~\cite{ioffe15}. Downsampling is performed via average pooling and upsampling  is performed via bilinear interpolation.

The code for the neural network training and inference was developed using the PyTorch~\cite{pytorch2019} library.

\subsection{Experiments}

We study the following variants of the Probabilistic U-Net by using different distributions (Section \ref{sec:contribution}) to model the latent space distributions(s):
\begin{itemize}
    \item Axis-aligned Probabilistic U-Net (AA)~\cite{kohl_probabilistic_2018}
    \item Full covariance Probabilistic U-Net (FC)
    \item Mixture AA Probabilistic U-Net
    \item Mixture FC Probabilistic U-Net
\end{itemize}

We used the \emph{generalized energy distance} (GED) metric~\cite{kohl_probabilistic_2018} to compare how well the distribution of neural network predictions $P_{s}$ matched the distribution of ground truth labels $P_{gt}$.

\begin{equation}
    D^2_{GED}(P_{gt}, P_s) =  2\mathbb{E}[d(S, Y)] - \mathbb{E}[d(S, S')] - \mathbb{E}[d(Y, Y')]
\label{eq:ged}
\end{equation}

The distance metric between a pair of segmentations, $d(x, y)$, is defined as $1 -\mathrm{IoU}(x, y)$~\cite{kohl_probabilistic_2018}, \cite{baumgartner_phiseg_2019}. In Equation \ref{eq:ged}, $S, S'$ are samples from the prediction distribution and $Y, Y'$ are samples from the ground truth distribution. The expectation terms in Equation \ref{eq:ged} are computed via Monte Carlo estimation. We used $16$ samples from the prediction distribution to compute the GED. A lower GED implies a better match between prediction and ground truth distributions. To check for statistical significance in the differences between the GED metric for models, we performed the Wilcoxon signed-rank test at a significance of $0.05$.

We also looked at the trends in the component terms of the GED metric. The first term $\mathbb{E}[d(S, Y)]$ signifies the extent of overlap between samples from the prediction and ground truth distribution, while the second term $\mathbb{E}[d(S, S')]$ is the sample diversity. By looking at trends in these terms in conjunction with the GED metric, we studied the interplay between overlap and diversity that lead to a better (or worse) match between distributions.

\begin{figure}[htb]
\centering
\includegraphics[width=0.8\linewidth]{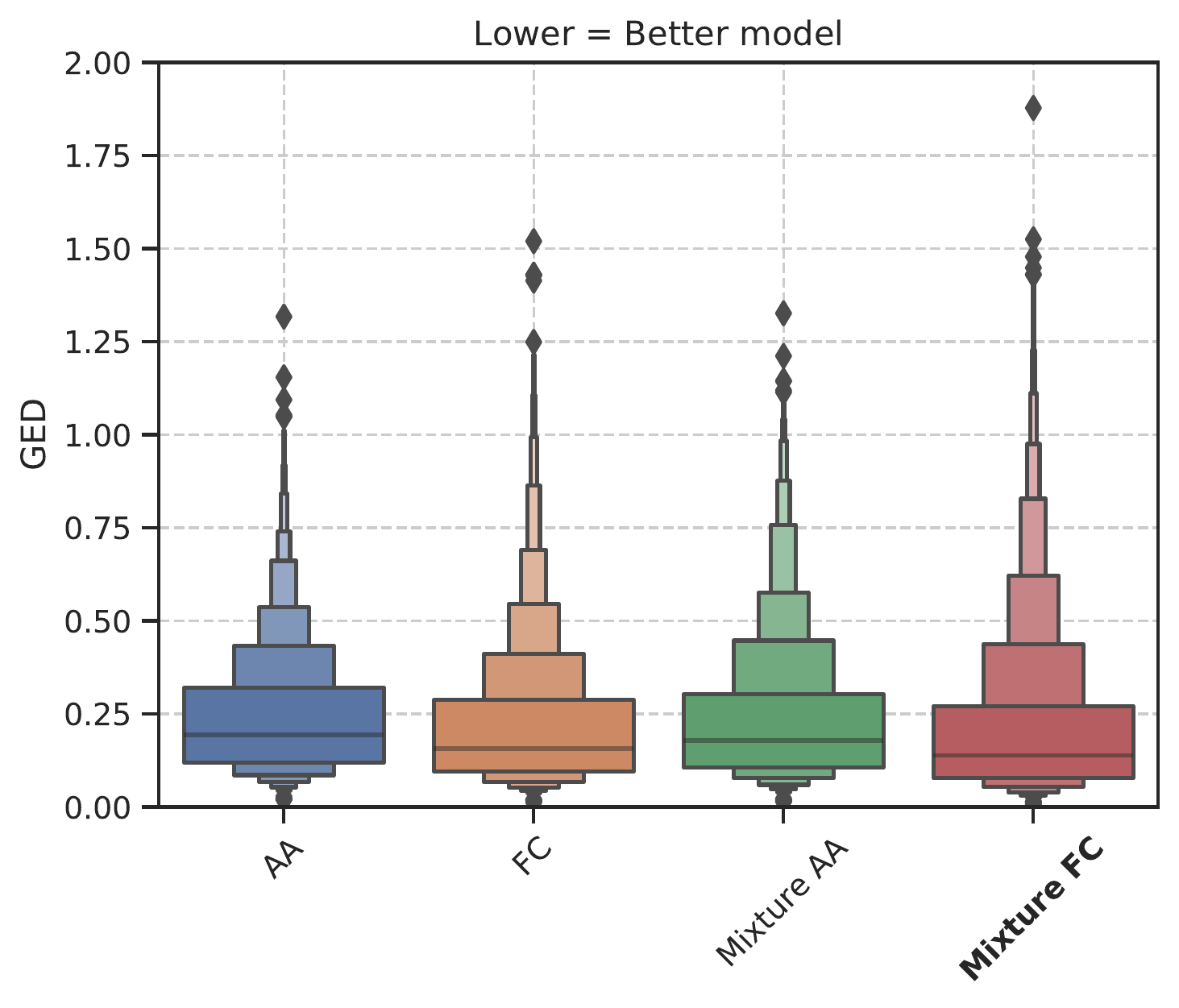}
\caption{Trends in the GED metric for LIDC-IDRI dataset using a letter-value plot. The median is shown by a horizontal line segment, and the innermost box is drawn at the upper and lower fourths, i.e. 25th and 75th percentiles. The other boxes are drawn at the upper and lower eighths, sixteenths and so on. The GED was computed using $16$ prediction samples. We found all pairwise differences to be statistically significant, with Mixutre FC to be the best performing model.}
\label{fig:ged_trends}
\end{figure}

\section{Results and Discussion}
\label{sec:rnd}

Figure \ref{fig:ged_trends} and Table \ref{tab:ged_stats} show the trend in the GED metric for the LIDC-IDRI dataset. We see that the Probabilistic U-Net using a mixture of full covariance Gaussians (Mixture FC) performs best w.r.t the GED metric. We also see that the models, using a full covariance Gaussian as the latent space distribution, did better than the standard Probabilistic U-Net (AA).
\begin{table}[h]
\centering
\begin{tabular}{@ {} l@{\hspace{1cm}} l@{\hspace{1cm}}}
\toprule
Model       & GED (mean $\pm$ std-dev) \\ \midrule
AA & $0.242 \pm 0.168$ \\
FC   & $0.220 \pm 0.188$  \\
Mixture AA       & $0.238 \pm 0.191$  \\
\textbf{Mixture FC} & $0.218 \pm 0.220$  \\
\bottomrule
\end{tabular}%
\caption{Mean and standard deviation of the GED metric for the LIDC-IDRI test-set.}
\label{tab:ged_stats}
\end{table}

\begin{figure}[htb]
\centering
\includegraphics[width=1.0\linewidth]{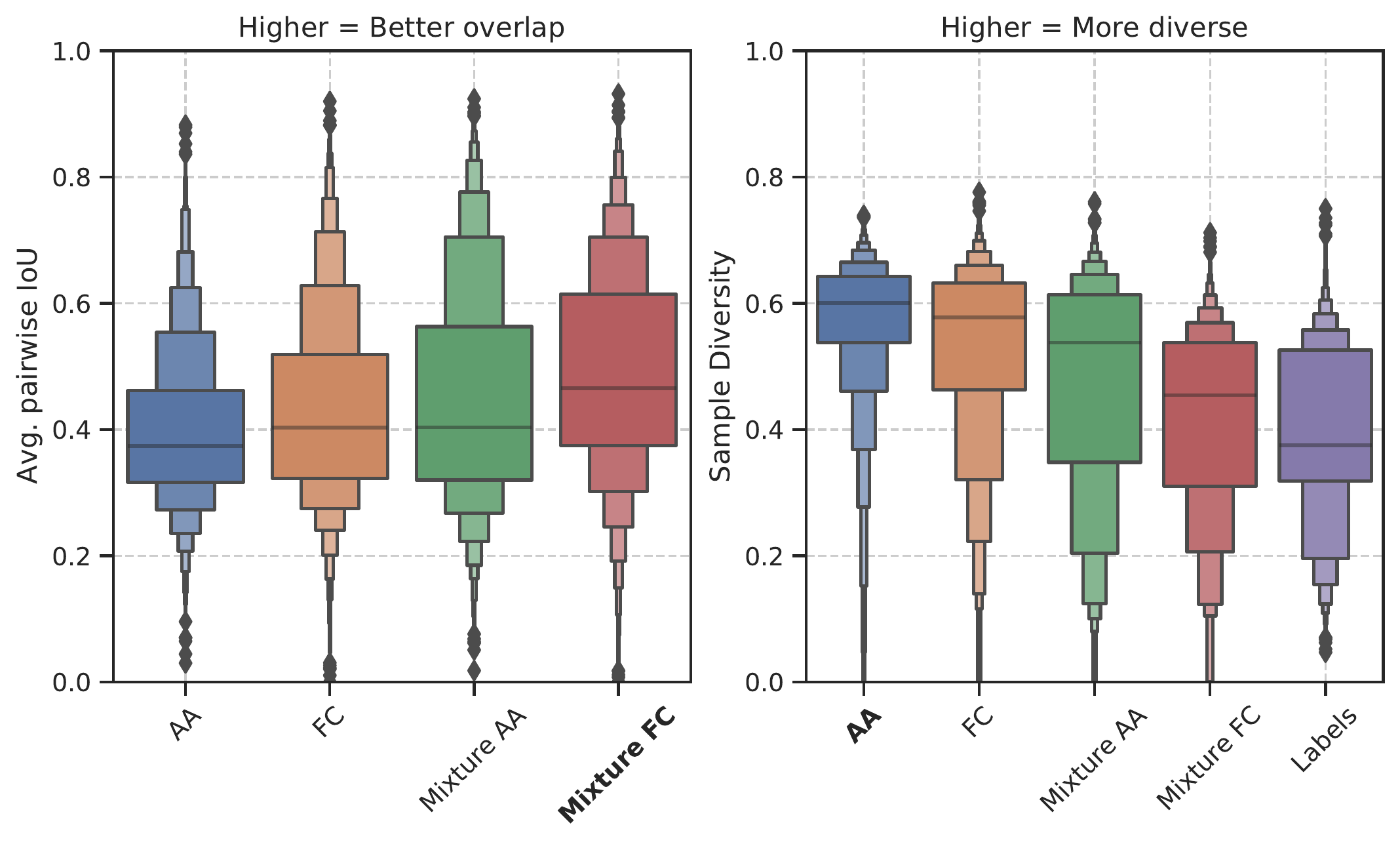}
\caption{GED break-up into average IoU and sample diversity terms for the LIDC-IDRI dataset. We observed that the trends in the GED metric were dictated by the tradeoff between overlap and diversity.}
\label{fig:ged_breakup}
\end{figure}

In Figure \ref{fig:ged_breakup}, our results show that the Mixture FC variant achieves an optimal combination between overlap and diversity to emerge as the best performing model with respect to the GED metric. The Mixture FC predictions matched the distribution of reference segmentations closest, with a higher overlap and a sample diversity closer to the inter-observer variability. We also observed that models using a Gaussian mixture distribution had better overlap but produced less diverse predictions than single Gaussian variants, for both axis-aligned and full covariance distributions. This has been illustrated visually with examples in Appendix \ref{sec:examples}.

A limitation in our generalized Probabilistic U-Net framework is the increase in the number of hyperparameters that need to be tuned for the distributions used. For example, mixture models need to find the optimal number of components ($N$), and the temperature ($\tau$) for the Gumbel-Softmax distribution, used to control the relaxation of the discrete categorical distribution. Therefore, finding the optimal model for a particular dataset could potentially be a time-consuming task. An interesting research direction could be treating the number of mixture components as a learned parameter, instead of it being fixed at the start of training.

\section{Conclusion}
Our work focused on extending the Probabilistic U-Net framework by going beyond the axis-aligned Gaussian distribution as the de facto choice for the variational distribution. Our results showed that the choice of model (or distribution) is dictated by the optimal overlap-diversity combination. Therefore, investigating the suitability of different latent space distributions is beneficial. The distributions studied in this paper can be used as a drop-in replacement for the axis-aligned Gaussian in other extensions to the Probabilistic U-Net framework, like the class of models using hierarchical latent spaces~\cite{kohl_hierarchical_2019}, \cite{baumgartner_phiseg_2019}, or as base distributions for normalizing flow based segmentation models~\cite{liu_uncertainty_2020}.

\section{Acknowledgements}
This work was financially supported by the project IMPACT (Intelligence based iMprovement of Personalized treatment And Clinical workflow supporT) in the framework of the EU research programme ITEA3 (Information Technology for European Advancement).

\bibliographystyle{splncs04}
\bibliography{refs}

\appendix
\section{Neural network hyperparameters}
\label{sec:hyperparams}
We used the Tune\cite{liaw2018tune} package to perform hyperparameters optimization. The number of hyperparameters changed based on the model/latent distribution. The number of U-Net encoder and decoder blocks was fixed at $3$, and the filter depths used were $32, 64, 128$ for the first, second and third blocks respectively. The bottleneck layer had a filter depth of 512. $\beta$ is the weight assigned to the KL-divergence term in the Probabilistic U-Net loss function. 

Table \ref{tab:hyper_search} contains the search space used to perform hyperparameter optimization.

\begin{table}[h]
\centering
\resizebox{0.8\textwidth}{!}{\begin{tabular}{@ {} l@{\hspace{1cm}} l@{\hspace{1cm}} l@{\hspace{1cm}} l@{\hspace{1cm}} l @ {}}
\toprule
Hyper-parameter       & \multicolumn{2}{c}{Search Space} \\ \midrule
              & Range    & Search type  \\
              \cmidrule{2-3}
Latent space dimension & $[2, 4, 6, 8]$ & Grid search \\
$\beta$   & $[1, 10, 100]$  & Grid search\\
Rank        & $[1, 5]$  & Random sampling\\
Mixture components & $[1, 10]$  & Random sampling \\
Temperature      & $[0.1, 0.5]$  & Random sampling \\
\bottomrule
\end{tabular}}%
\caption{Hyperparameter search space}
\label{tab:hyper_search}
\end{table}

\begin{table*}[htp]
\centering
\resizebox{1.0\textwidth}{!}{\begin{tabular}{@ {} l@{\hspace{1cm}} l@{\hspace{1cm}} l@{\hspace{1cm}} l@{\hspace{1cm}} l @ {\hspace{1cm}} l @ {\hspace{1cm}}}
\toprule
Model       & \multicolumn{5}{c}{hyperparameters} \\ \midrule
              & Latent space dimension    & $\beta$ & Rank & Mixture components & Temperature \\
              \cmidrule{2-6}
AA & 6 & 1 & - & - & - \\
FC   & 2 & 1 & - & - & - \\
AA Mixture  & 4 & 1 & - & 5 & 0.36 \\
FC Mixture      & 2 & 1 & - & 9 & 0.28\\
\bottomrule
\end{tabular}}
\caption{LIDC-IDRI hyperparameters}
\label{tab:hyper_lidc}
\end{table*}

\section{Example predictions}
\label{sec:examples}

\begin{figure*}[!htbp]
\centering
\begin{subfigure}[t]{0.8\linewidth}
\includegraphics[width=1.0\linewidth]{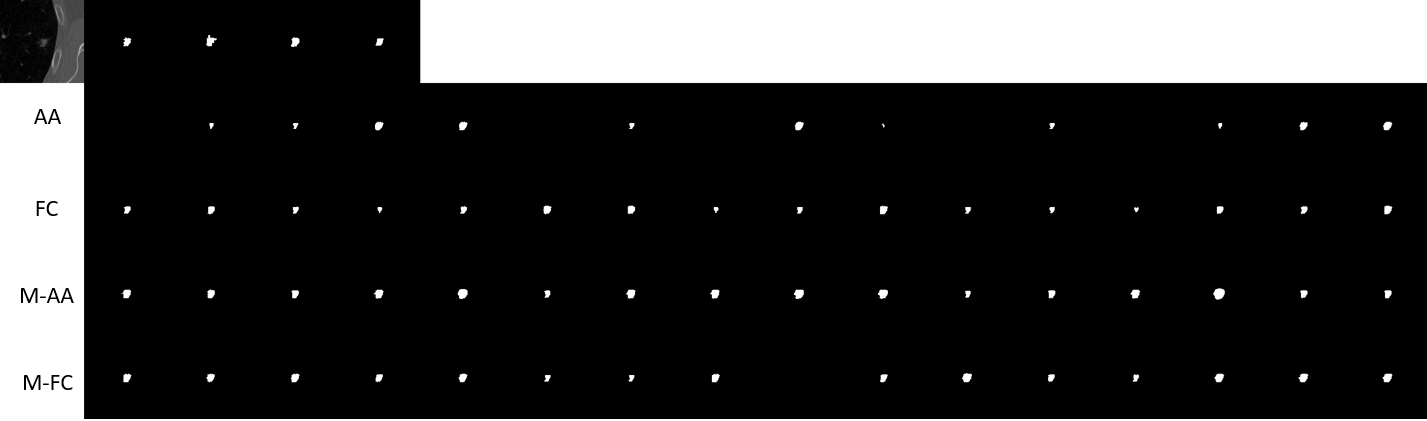}
  \caption{This example shows the reduced sample diversity (and improved overlap) of the mixture models (M-AA, M-FC) with respect to the full-covariance (FC) and axis-aligned (AA) Gaussians. Specifically, the AA predictions are diverse but do not match the distribution of the reference segmentations (poor overlap).}
  \label{fig:lidc_tile_1}
\end{subfigure}
\begin{subfigure}[t]{0.8\linewidth}
  \includegraphics[width=1.0\linewidth]{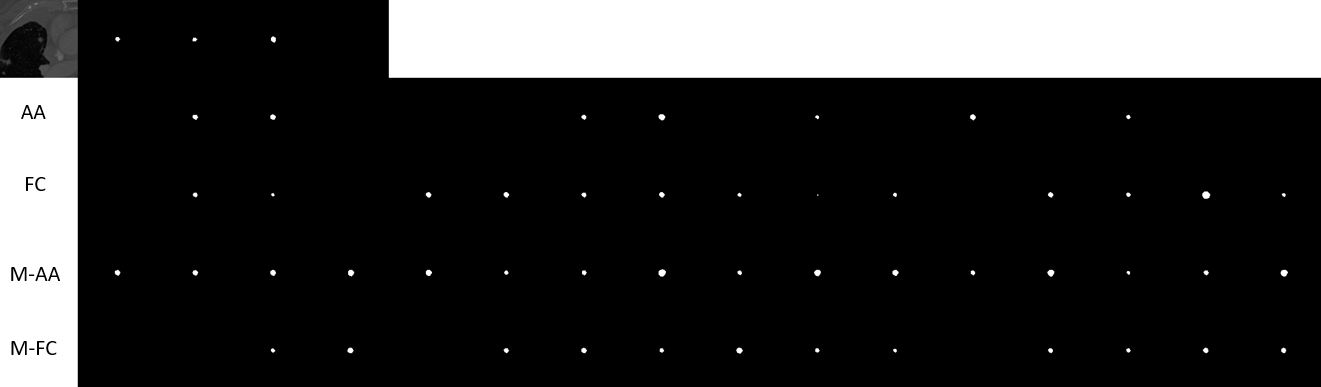}
  \caption{This example shows an instance where the Mixture FC (M-FC) and FC predictions perfectly match the distribution of the reference segmentations, with $75$\% of its predictions containing a lesion.}
  \label{fig:lidc_tile_2}
\end{subfigure}
\caption{Example image, labels, and predictions for the LIDC-IDRI dataset. The first row contains the image and the reference segmentations. The following rows show $16$ samples drawn from the prediction distribution (used to compute the GED).}
\label{fig:image_tiles}
\end{figure*}

To support the quantitative results in Section \ref{sec:rnd}, we present example predictions, along with images and reference segmentations, in Figure \ref{fig:image_tiles}.

\end{document}